\pdfoutput=1

\documentclass[11pt]{article}

\usepackage{todonotes}

\usepackage{acl}

\usepackage{times}
\usepackage{latexsym}

\usepackage{pifont}

\usepackage[T1]{fontenc}

\usepackage[utf8]{inputenc}

\usepackage{microtype}

\usepackage{multirow}
\usepackage{comment}
\usepackage{wrapfig}
\usepackage{subcaption}
\usepackage{amsmath,amsthm,amssymb}
\usepackage{graphicx}
\usepackage{xcolor}

\usepackage{enumitem}
\usepackage{makecell}

\newcommand\SYSNAME{MLP}

\newcommand{\bluebox}[1]{\setlength{\fboxsep}{1pt}\colorbox{blue!20}{#1}}

\newcommand{\stt}{\footnotesize \tt}

\usepackage{cleveref}
\crefname{section}{s}{ss}
\Crefname{section}{s}{ss}
\crefname{table}{Table}{}
\crefname{figure}{Fig.}{}
\crefname{algorithm}{Alg.}{}
\crefname{equation}{Eq.}{}
\crefname{appendix}{App.}{}
\crefformat{section}{\S#2#1#3} 

\newcommand{\hide}[1]{}

\title{Jointly Identifying and Fixing Inconsistent Readings\\from Information Extraction Systems\\
}

\author{
  {Ankur Padia\thanks{*This work was done while the first author was doing his Ph.D. at the University of Maryland, Baltimore County and before joining Philips Research North America.}}\ , Francis Ferraro and Tim Finin \\
  University of Maryland, Baltimore County\\
  Baltimore, MD 21250 USA\\
  \texttt{\{pankur1, ferraro, finin\}@umbc.edu} \\}

\begin{document}
\maketitle
\begin{abstract}

Information extraction systems analyze text to produce entities and beliefs, but their output often has errors. In this paper we analyze the reading \textit{consistency} of the extracted facts with respect to the text from which they were derived and show how to detect and correct errors. We consider both the scenario when the provenance text is automatically found by an IE system and when it is curated by humans. We contrast \textit{consistency} with \textit{credibility}; define and explore \textit{consistency and repair tasks}; and demonstrate a simple, yet effective and generalizable, model. We analyze these tasks and evaluate this approach on three datasets.
Against a strong baseline model, we consistently improve both consistency and repair across three datasets using a simple \SYSNAME{} model with attention and lexical features.

\end{abstract}

\section{Introduction}
\label{sec:introduction}

Information Extraction (IE) systems \nocite{nell2015cmu,finin2017hltcoe,ernst2014knowlife,suchanek2007yago,surdeanu2010simple} read text to extract entities, and relations and create beliefs represented in a knowledge graph. Current systems though are far from perfect: e.g., in the 2017 Text Analysis Conference (TAC) Knowledge Base Population task, participants created knowledge graphs with relations like \textit{cause of death} and \textit{city of headquarters} from news corpora \cite{tac2017}. When manually evaluated, no system had achieved an F1 score above 0.3 \cite{TAC17slides}.

One reason for such low scores is \textit{inconsistency} between the text and the extracted beliefs. We consider a belief to be \textit{consistent} if the text from which it was extracted linguistically supports it (regardless of any logical or real-world factual truth). We show the difference between consistent and inconsistent readings, along with a potential correction, in \cref{fig:example}. In \cref{fig:inconsistentNoFix}, the system considered {\small \tt Harry Reid} was charged with an {\small \tt assault}, which is not consistent with the provenance sentence. In \cref{fig:consistent} the system is consistent in constructing its belief. %

\begin{figure}[!ht] \small 
	\centering 
	\begin{subfigure}{\linewidth}
		\begin{tabular}{p{0.9\linewidth}}
			\textbf{Belief learned by IE system:} \\

			\addtolength{\leftskip}{3mm}{\footnotesize \tt
				per:charges(Harry Reid, \underline{assault})} \smallskip\\

			\textbf{Provenance identified by IE system:}\\
			\addtolength{\leftskip}{3mm}{\small Nevada's \bluebox{Harry Reid} switches longtime stance to support \bluebox{assault} weapon ban} \smallskip\\
			\textbf{Analysis output:}\\
			\addtolength{\leftskip}{3mm}{\small Is reading consistent: \quad Inconsistent}\\ 

			\addtolength{\leftskip}{3mm}{\small Suggested relation:\quad\quad no repair}\\

		\end{tabular}
		\caption{An inconsistent reading with no correction.}
		\label{fig:inconsistentNoFix}
	\end{subfigure}

        \vspace{0.5em}
	\begin{subfigure}{\linewidth}
		\begin{tabular}{p{0.9\linewidth}}
			\vspace{-0.2cm}
			\textbf{Belief learned by IE system:} \\
			\addtolength{\leftskip}{3mm}{\footnotesize \tt
				per:cause\_of\_death(Edward Hardman, \underline{Typhoid fever})} \smallskip\\
			\textbf{Provenance identified by IE system:}\\
			\addtolength{\leftskip}{3mm}{\small The Western Australian government agreed to offer the Government Geologist post to \bluebox{Hardman} shortly before news of his death reached them . Early in April , he contracted \bluebox{\underline{typhoid fever}} , and died a few days later in a Dublin hospital on 6 April} \smallskip\\
			\textbf{Analysis output:}\\

			\addtolength{\leftskip}{3mm}{\small Is reading consistent:\quad Consistent}\\ 
			\addtolength{\leftskip}{3mm}{\small Suggested relation: \ \ \quad per:cause\_of\_death}
		\end{tabular}
		\caption{A consistent reading not requiring a correction. Notice the relation is unchanged.}
		\label{fig:consistent}
	\end{subfigure}
	\caption{Examples of beliefs extracted from real IE systems on the TAC 2015 English news corpus, demonstrating the \textit{consistency} and \textit{repair} tasks. Multiple sentences can contribute to a belief~(\ref{fig:consistent}).}
        \label{fig:example}
\end{figure}

We study two problems: (i) whether an extracted belief is consistent with its text (called consistency), and (ii) correcting it if not (called repair). We believe we are the first to study these problems jointly. We model these problems jointly, arguing that addressing both of these is important and can benefit one another. Our use of \textit{consistency} here refers to a language-based sense that text supports the belief even if its contradicts world knowledge.
We are concerned with methods that can be \textit{standalone}---that is, reliant on neither a precise schema~\cite{KGEval} nor an ensemble of IE systems, e.g., \citet{yu2014wisdom,viswanathan2015stacked}. %
Previous work on determining the consistency of an IE extraction was not standalone. We want a standalone approach because the results from non-standalone approaches cannot be applied when only the beliefs and associated provenance text is available without the IE ensemble systems and schema. (For this study we consider English beliefs and provenance sentences.) Parallel to the broad IE domain, schema-free and standalone systems have been developed to verify the credibility of news claims \cite{popatDeclare,riedel2017simple,rashkinFake}, but we are not aware of a study of their performance on IE system tasks. We incorporate these credibility systems into our study in order to determine their applicability for our tasks. 
We make the following contributions.

\begin{description}[leftmargin=2mm,itemsep=0.25mm]
\item[A study of real IE inconsistencies.] 
We catalog and examine the understudied aspect of language-based consistency (\cref{sec:catelogingIEsystemerrors}).


\item[A novel framework.] To our knowledge we are the first to study and propose a framework for joint consistency and repair (\cref{sec:approach}). 

\item[Analysis of techniques.] We show the effectiveness of straightforward techniques 
compared to more complicated approaches (\cref{sec:experiments}). 

\item[Study of different provenance settings.]
We consider and contrast cases where provenance sentences are retrieved by an IE system (as in TAC) vs. where they are curated by humans (as in \newcite[TACRED]{zhang2017position}). 


\end{description}

\section{Task Setup}

\subsection{Consistency and Repair}
We say the belief was consistently read if the text \textit{lexically} supports the belief. While this can be viewed as a lexical entailment, it is not a logical, causal, or broader inferential/knowledge entailment. 
For example the belief {\small \tt{<Barack Obama,per:president\_of,Kenya>}} is consistent with a provenance sentence {\small \tt{``Barack Obama, president of Kenya, visited the U.S. for talks}"} even though the sentence falsely claims that Obama is president of Kenya. .

The belief is considered repaired if the relation extracted by the IE system was not supported by the text, but when replaced by another relation that is supported by the text.

\subsection{Datasets}
\label{sec:experiments:data}

We use three datasets: TAC 2015, TAC 2017, and a novel dataset we call TACRED-KG. %
All datasets use actual output from real IE systems. 
Each dataset is split into train/dev/test splits: in \cref{tab:dataset_stats} (in the appendix) we show the size of each split, in terms of the number of provenance-backed beliefs.

\paragraph{TAC 2015 and 2017.} These include the output of 70+ IE systems, from the TAC 2015 and TAC 2017 shared tasks, with belief triples supported by up to four provenance sentences. Each belief was evaluated by an LDC expert \cite{kbp15assessment}. We used these LDC judgments as the consistency labels for our experiments. For TAC 2015, 27\% of the 34k beliefs are judged consistent; for TAC 2017, 36\% of the 57k beliefs are judged consistent. %

These TAC datasets do not, however, contain information on possible corrections when the belief is inconsistent.
To overcome this limitation, we used negative sampling on the consistent beliefs with their provenance to create an inconsistent pair. We first selected an entity and then identified a set of relations that apply to the entity. We randomly chose one of the relations with uniform probability and shuffled it with another relation, keeping the provenance the same. For example, given two consistent beliefs {\stt Barack\_Obama,president\_of,US}, and {\stt Barack\_Obama,school\_attended,Harvard}, we swap {\stt president\_of} with {\stt school\_\-attended}, keeping the provenance unchanged. This yields inconsistent beliefs associated with corresponding provenance and the correct labels.

\begin{table*}
	\centering
	\small
        \resizebox{\textwidth}{!}{
		\begin{tabular}{|p{1.1cm}|p{3.8cm}|p{10.55cm}|}                
			\hline
			\textbf{Category} & \textbf{Definition} & \textbf{Extracted Belief followed by IE extracted provenance text} \\
                        \hline
                        Incorrect & subject \& object present but  &\textit{Harry Reid   \qquad   per:charges    \qquad   assault}  \\
			relation & relation not triggered/entailed & Nevada's \bluebox{Harry Reid} switches longtime stance to support \bluebox{assault} weapon ban\\
                        
			\hline
			Subject & entity is not mentioned in  & \textit{Eleanor Catton	 \qquad gpe:subsidiaries \qquad	Bain} \\
			missing & provenance & Buying into Canada Goose is the latest Canadian investment for \bluebox{Bain}.\\
			\hline
			Misc & fact does not adhere to &\textit{Reginald Wayne Miller     \qquad        per:charges         \qquad   felony}  \\
			& schema-specific guidelines and requirements & Various news outlets have reported that federal agents have probable cause to charge \bluebox{Reginald Wayne} \bluebox{Miller} with forced labor, a \bluebox{felony} that can carry up to a twenty-year prison sentence per charge.\\
			\hline
			Object &entity is not mentioned in  &\textit{Kermit Gosnell    \qquad per:cities\_of\_residence   \qquad  America } \\
			missing & provenance & Historic crowdfunding for movie about abortionist \bluebox{Kermit Gosnell} - YouTube \\
			\hline
		\end{tabular}%
	}
	\caption{Examples for each of the four identified error categories from the TAC 2015 dataset.}
	\label{tab:tac-error-types}%
\end{table*} %

\paragraph{TACRED-KG.}
The TACRED-KG dataset is a novel adaptation from the existing TACRED \cite{zhang2017position} relation extraction dataset. %
TACRED is focused on providing data for typical relation extraction systems. As such, it contains 4-tuples (subject, object, provenance sentence, correct relation), where relation extraction systems are expected to predict that relation for the given subject-object pair and the sentence. We turn this relation extraction dataset into a KG-focused dataset. 
We then used a relation extraction position-aware attention RNN model \cite{zhang2017position}  system on the TACRED data to produce 5-tuples (subject, object, provenance sentence, correct relation, predicted relation). From these we created a provenance-backed KG dataset, TACRED-KG, as (subject, predicted relation, object, provenance sentence). 
In TACRED-KG, we treat the gold standard relation as the repair label. We consider beliefs consistent when the predicted and gold standard relations are the same. %

\paragraph{Observational Comparison.}
We note some qualitative observations about these datasets, though traceable back to how each dataset was constructed. First, TAC 2015 and TAC 2017 contain more provenance examples \textit{per relation} than TACRED-KG. Second, because the provenance was provided by varied IE systems in TAC 2015/2017, the provenance may be the result of noisy extractions and matching: the provenance for TAC 2015/2017 is often noisier than TACRED-KG (e.g., portions of sentences vs. full sentences).

\section{What Errors Do IE Systems Make?}
\label{sec:catelogingIEsystemerrors}

We begin with an analysis of errors in the beliefs from actual IE systems. 
This analysis is enlightening, as each system used different approaches and types of resources to extract potential facts. 

We sampled 600 beliefs and their provenance text each from the training portions of three different knowledge graph datasets: TAC 2015, TAC 2017, and TACRED-KG. As described in \cref{sec:experiments:data}, they all contain provenance-backed beliefs that were extracted from actual IE systems (but ones which are generally not available for subsequent downstream examination). All of the beliefs are represented as a relation between two arguments. The authors manually assessed these according to available and published guidelines~\cite{kbp15assessment,kbp15slots,tac2017} to understand the kinds of errors made by the IE systems. We identified four types of errors: the subject (first argument) not present in the provenance text; the object (second argument) not present in the provenance; an insufficiently supported relation between two present arguments; and relations that run afoul of formatting requirements, e.g., misformed dates. 
We show examples of these in \cref{tab:tac-error-types}.



\begin{figure}[t]
\centering
  \includegraphics[width=0.45\textwidth]{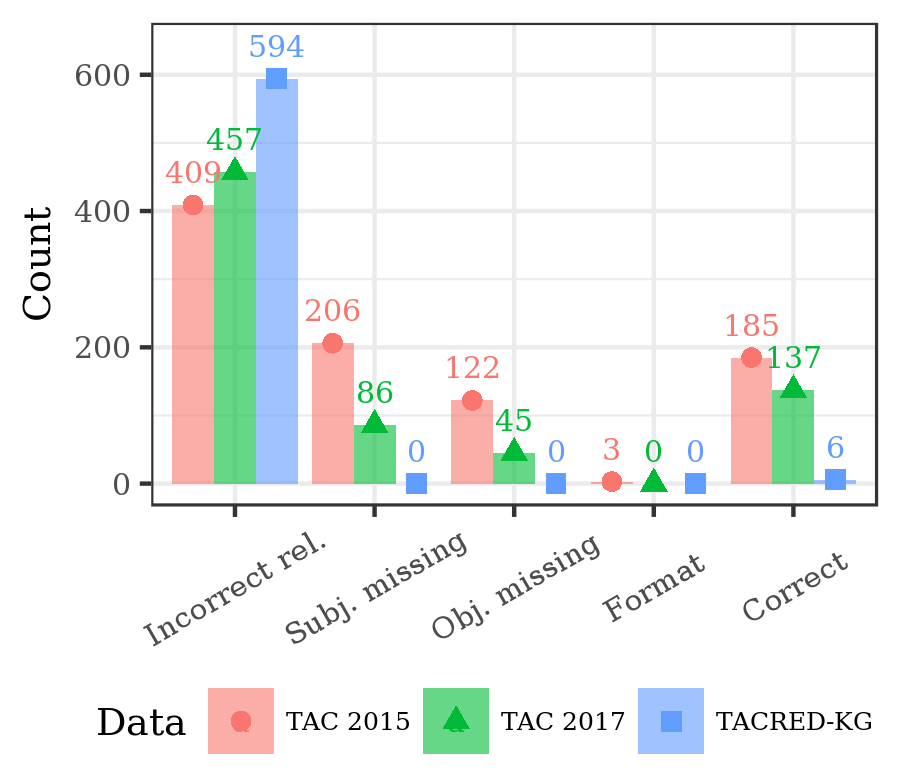}
  \caption{Error categorization of 600 beliefs extracted by IE systems on three datasets. Multiple categories can apply as beliefs can have incorrect relations and incomplete provenance.}
  \label{fig:previous-tac-errors}
\end{figure}

Our analysis, summarized in \cref{fig:previous-tac-errors}, found that the most frequent error type is an incorrect relation, followed by missing subject, missing object and (at a trace level) formatting errors. Though it varied based on dataset, approximately two-thirds of the sampled belief-provenance pairs had errors. %
The prevalence of incorrect relations \textbf{motivates the importance of the relation repair task}. %
It should be noted that while TAC 2015 and 2017 have a number of instances of missing subjects and objects, this is not the case for TACRED-KG. This illustrates a fundamental difference in selecting provenance information manually vs. automatically, and one that we observe to be experimentally important (\cref{sec:results}), between TAC 2015/2017 and TACRED-KG.

\section{Approach}
\begin{figure*}[t!]
        \centering
	\includegraphics[scale=.50]{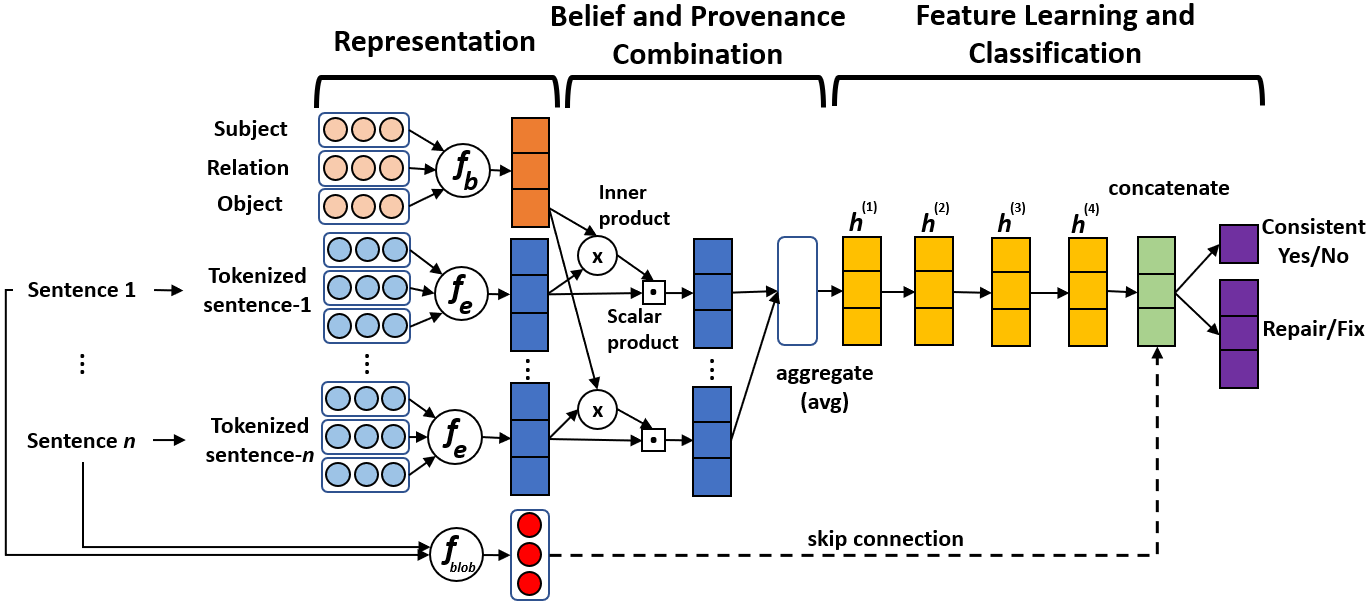}
	\caption{Given a belief and a set of $n$ provenance sentences,
          our framework determines its consistency and suggests a
          repair when if is deemed inconsistent. Our approach has
          three main modules: representation (\ref{sec:approach:representation}), combination (\ref{sec:approach:combination}), and feature learning and classification (\ref{sec:approach:features}).}
	\label{fig:crmtarch}
\end{figure*}

\label{sec:approach}



Our approach computes both the consistency of a belief $b_i$ and a ``repaired'' belief with respect to a given set of provenance sentences. We represent $b_i$ as a triple $\langle \textrm{subject}_i, \textrm{predicate}_i, \textrm{object}_i\rangle$ and the set of provenance sentences as $S_{i,1}, S_{i,2},...S_{i,n}$. The system outputs two discrete predictions: (1) a binary one indicating whether the belief is consistent with the sentences, and (2) a categorical one suggesting a repair. 
\Cref{fig:crmtarch} illustrates our approach for representing and combining the beliefs and provenance sentences to jointly learn the two tasks. %

Our approach has three main steps: embedding a belief and its provenance sentences in a vector space (\cref{sec:approach:representation}), combining/aggregating these representations (\cref{sec:approach:combination}), and using the result for additional feature learning and classification (\cref{sec:approach:features}). We describe our loss objective in \cref{sec:approach:optimization}. %
As we show, our framework can be thought of as generalizing high performing credibility models, such as DeClarE \cite{popatDeclare} or LSTM-text \cite{rashkinFake}.

\subsection{Belief \& Provenance Representation}
\label{sec:approach:representation}

We process and tokenize a belief's arguments and relation. For example, the belief $\langle{\tt Barack\_Obama}$, ${\tt per:president\_of}$, ${\tt United\_States}\rangle$ yields a subject span (``Barack Obama''), a relation span (``president of''), and an object span (``United States''). %
We input processed text through an embedding function $f_{\textit{belief}}$ to get a single embedding $\boldsymbol{b}$ for the belief. Here, $f_{\textit{belief}}$ could be average of pre-trained word embeddings, or  final hidden state obtained from a sequence model (LSTM or Bi-LSTM) or the embedding from a transformer model (e.g., BERT~\cite{devlin2019bert}). As we discuss in \cref{sec:experiments:components}, we experiment with all of these.

We represent the provenance sentences at two granularities. The first is by representing each sentence separately. We get a representation $\boldsymbol{s}_i$ for each provenance sentence via an embedding function $f_{\textit{evidence}}$ that embeds and combines them into a single vector. 
We define $f_{\textit{evidence}}$ similarly to $f_{\textit{belief}}$. 

The second level considers all sentences at the same time. We refer to this as \textit{blob}-level processing (rather than paragraph- or document-level) since the provenance sentences may come from different documents and we cannot assume any syntactic continuity between sentences. We obtain a representation of the blob from $f_{\textit{blob}}$. In principle any method of distilling potentially disjoint text could be used here: we found TF-IDF to be effective, especially as multiple sentences of provenance selectively extracted from different sources could result in lengthy, but non-narratively coherent text (which can be problematic for transformer models). 




\subsection{Belief and Provenance Combination}
\label{sec:approach:combination}

Given the belief and provenance representations, we compute their similarity $\alpha_i$ as the cosine of the angle between their embedded representations: $\alpha_{i} = \frac{\boldsymbol{b_{i}^{T}s_{i}}}{||\boldsymbol{b_{i}}||\cdot||\boldsymbol{s}_{i}||}$. The intuition is that sentences that are more consistent with the belief will score higher than those which are less.
Scoring is important, as each IE system may give multiple provenance sentences (e.g., TAC allowed four).  The sentences can be correct and support the belief, or be poorly selected and unsupportive. Higher scores suggest the provenance is related to the belief and helps differentiate supportive from unsupportive provenance.
We use the computed similarity scores to combine the provenance representations and take a weighted average as our final input, capturing the semantics of the belief and provenance, as $\boldsymbol{x} = \frac{1}{n}\sum_{i}\alpha_{i}\cdot\boldsymbol{s}_{i}$.
We pass the created representation $\boldsymbol{x}$ as the input to the feature learning module. 

Though our computation of $\mathbf{\alpha}_i$ and $\mathbf{x}$ operate at the sentence-level, our approach can also be applied to individual word representations. For this word-level attention, we replace each sentence representation $s_{i}$ with a word representation $w_{ij}$ in our computation of $\mathbf{\alpha}_i$ and $\mathbf{x}$. While we experimented with this word-level attention we found the model had trouble learning, frequently classifying beliefs nearly all as consistent, or inconsistent with ``no repair.'' We note that a similarly effective word-level attention was provided in DeClarE. 

We selected a similarity-based, rather than position-based, attention. Applying position-based attention, as \newcite{zhang2017position} did on the TACRED dataset, assumes that provenance sentences contain an explicit mention of the subject and object. In our setting that explicitly is not the case (recall the prevalence of missing arguments in our datasets, c.f. \cref{fig:previous-tac-errors}). There is also an assumption that there is exactly one provenance sentence as opposed to TAC, where an IE system can select up to four provenance sentences without explicitly mentioning either the subject or object. %
\subsection{Feature Learning and Classification}
\label{sec:approach:features}
Prior to classification we may learn a more targeted representation $\mathbf{z}$ by, e.g., passing the combined representation $\boldsymbol{x}$ into a multi-layer perception. If we do not, then the consistency and repair classifiers operate directly on $\mathbf{z} = \mathbf{x}$.

We noticed through development set experiments that while adding additional layers initially helped, using more than three layers marginally decreased performance. For a $k$-layer MLP we obtained the projections $\boldsymbol{h}^{(j)}$, for $1\leq j\leq k$, as:
$\textbf{h}^{(j)} = g\left(\boldsymbol{W}^{(j)}\boldsymbol{h}^{(j-1)}+\boldsymbol{b}^{(j)}\right).$
$\boldsymbol{h}^{(0)} = \boldsymbol{x}$ indicates the input, $\boldsymbol{W}^{(j)}$ and $\boldsymbol{b}^{(j)}$ are each layer's learned weights and biases (respectively), and $g$ is the activation function. Through dev set experimentation we set $g$ to be ReLU \cite{glorot2011deep}. We found the MLP gave better performance (\cref{sec:experiments}) and that it was parametrically and computationally efficient. We note that the effectiveness of an MLP was also noted by the two top systems from the Fake News Challenge~\cite{fncWinnerRank2,riedel2017fnc} for the verification task. On dev, we evaluated from one to five hidden layers and found the performance to be consistent after three layers, with the mean close the scores in Tables \ref{tab:consistencyPerformance} and \ref{tab:repairPerformance} and a maximum standard deviation across all the dataset and evaluation metrics to be less then one F1 point.


In addition to the learned features learned $\textbf{h}^{(k)}$, we experiment with a lexically-based skip connection, where the input from the previous layer skips a few layers and is connected to a deeper one. We found this to be effective when making use of ``blob'' level features, computed via $f_{\textit{blob}}$. 
We further found computing $f_{\textit{blob}}$ as the TF-IDF vector of all provenance text to be especially effective (\cref{sec:ablationstudy}). When using this connection, we compute $\mathbf{z} = \left[\boldsymbol{h}^{(k)}, f_{\textit{blob}}(blob) \right]$. %
If this connection is not used, $\mathbf{z} = \boldsymbol{h}^{(k)}$. %
\paragraph{Classification.}
We use the final representation $\mathbf{z}$ as input to the consistency ($\hat{y}_c = {\tt sigmoid}\left(\boldsymbol{W}^{c}\boldsymbol{z}+\boldsymbol{b}^{c}\right)$) and repair classifiers ($\hat{\boldsymbol{y}}_{r} = {\tt softmax}\left(\boldsymbol{W}^{r}\boldsymbol{z}+\boldsymbol{b}^{r}\right)$). The parameters $W^{c}$ and $W^{r}$ have sizes $1 \times (d_{\text{tf-idf}}+d_{\text{hidden}})$ and $d_{\text{relations}} \times (d_{\text{tf-idf}}+d_{\text{hidden}})$, respectively. Here $d_{\text{tf-idf}}, d_{\text{hidden}}$, and $d_{\text{relations}}$ are the dimension of the TF-IDF vector, hidden vector and number of relations considered by the IE systems.



\subsection{Joint Optimization}
\label{sec:approach:optimization}
We train the parameters using back propagation of both losses, $\mathcal{L}_{consistency}$ and $\mathcal{L}_{repair}$, jointly:
\begin{equation} 
\mathcal{L}  =
\mathcal{L}_{\textrm{consistency}}\left(y_{c},\hat{y}_{c}\right)+
\mathcal{L}_{\textrm{repair}}\left(\boldsymbol{y}_{r},\boldsymbol{\hat{y}}_{r}\right)
\label{eqn:joint-loss}
\end{equation} 
Each subloss is a cross-entropy loss between the true ($y_c, \boldsymbol{y}_r$) and predicted ($\hat{y}_c, \boldsymbol{\hat{y}}_r$) responses, weighted inversely proportional to the prevalence of the correct label.
The tasks are not independent. In our formulation they share the same provenance and belief representations so learning both tasks jointly helps in learning these shared parameters.\footnote{See \cref{sec:experiments} for discussion of alternative losses.}

While in this paper we present a joint loss objective, we note that we separately experimented with alternative, non-joint approaches to \cref{eqn:joint-loss}. However, in development we found they performed worse than the joint approach. First we evaluated pipelined approaches, e.g., where the repair classifier also considered the output of the credibility model, but found its performance to be inferior to the joint approach. 
Second, we also tried using the repair output as input to the credibility classifier, and found that it resulted in high recall with poor precision, with inconsistent instances being classified as consistent. 
The shared abstract representation of belief and provenance used in our formulation presented above allows fine tuning for both subtasks. We also experimented on dev with other types of weighting, such as a uniform weighting. However, the inversely proportional weighting scheme we describe in the main paper is what performed best on dev experiments.

\paragraph{A Generalizing Framework.}
We note that we can represent DeClarE by defining the belief encoder $f_{\textit{belief}}$ as averaging word embeddings, a provenance encoder $f_{\textit{evidence}}$ to be a Bi-LSTM, combining these representations with word level attention, and passing them to a two layer MLP without lexical skip connections. %
To achieve this specialization, we can optimize either $\mathcal{L}_{\textrm{consistency}}$ or 
$\mathcal{L}_{\textrm{repair}}$. %
Representing LSTM-text is similar.
This shows that our framework encompasses prior work. %

\section{Experiments}
\label{sec:experiments}

We centered our study around four questions, answered throughout \cref{sec:results}. \textbf{(1)} As our approach subsumes credibility models, can those credibility models also be used for the consistency and/or repair tasks (\cref{sec:credibilitymodel})? \textbf{(2)} What features and representations are important for the consistency and repair tasks (\cref{sec:features})? \textbf{(3)} How important is it to model the realized (sequential) order of words within the provenance sentences for our tasks (\cref{sec:sequential})? \textbf{(4)} What are the differences between relation repair and extraction (\cref{sec:relationExtraction})? 

\subsection{Datasets and Hyperparameter Tuning}
Table \ref{tab:dataset_stats} provides statistics on the train/dev/test splits.
\begin{table}[t]
\small
  \centering
    \begin{tabular}{lccc}
          & \multicolumn{1}{l}{TAC 2015} & \multicolumn{1}{l}{TAC 2017} & \multicolumn{1}{l}{TACRED-KG} \\
    Train & 20575 & 45841 & 68124 \\
    Dev   & 6859  & 5734  & 22631 \\
    Test  & 6856  & 5729  & 15509 \\
    \end{tabular}%
      \caption{Dataset statistics, in the number of provenance-backed beliefs, for the train/dev/test splits per dataset.}
  \label{tab:dataset_stats}%
\end{table}%
On dev, we tuned hyper-parameters over all the models and datasets, using learning rates from $\lbrace$10$^{-1}$,...,10$^{-5}\rbrace$ by powers of 10, dropout~\cite{srivastava2014dropout} from $\lbrace0.0,0.2\rbrace$, and L2 regularizing penalty values from $\lbrace0.0, 0.1...,0.0001\rbrace$ (powers of 10). We ran each model until convergence or for 20 epochs (whichever came first) with a batch size of 64. %

\begin{table*}[t]
	\centering
	\resizebox{0.99\textwidth}{!}{
	\begin{tabular}{|c|c|c|c|c||c|c|c||c|c|c||c|c|c|}
		\hline
		\textbf{$f_{belief}$} & \textbf{$f_{evidence}$} & \textbf{Comb.} & \textbf{Feat.} & \textbf{Sparse}& \multicolumn{3}{c||}{\textbf{TACRED-KG}} & \multicolumn{3}{c||}{\textbf{TAC-17}} & \multicolumn{3}{c|}{\textbf{TAC-15}} \\
		\hline
		& & & & & \textbf{P} & \textbf{R} & \textbf{F1} & \textbf{P} & \textbf{R} & \textbf{F1} & \textbf{P} & \textbf{R} & \textbf{F1} \\
		\hline\hline
		None & None & No & None & Binary & 63.96 & 83.46 & 72.42 & 19.65 & 5.29 & 8.34  &  28.08 & 0.81  & 1.58 \\
		None & None & No & None & TF-IDF & 63.95 & \textbf{83.24} & \textbf{72.33} & \textbf{57.58} & 30.66 & 14.05 & 22.68 & 15.08 & 18.12 \\
		\hline
		
		\hline\hline
		None & $\spadesuit$   LSTM & No & MLP & No & 42.59 & 66.66 & 51.98 & 52.05 &  30.76& 27.78 & 17.01 & 9.21 & 11.95 \\
		\hline
		BoW & $\clubsuit$  Bi-LSTM & Yes (W) & MLP & No & 42.59 & 66.66 & 51.98 & 37.31 & 52.44 & 43.54 & 31.17 & 36.55 & 33.65 \\
		\hline \hline
		BERT &  BERT  & Yes (S) & MLP & TF-IDF & \textbf{66.42}  & 76.26   & 69.99   & 48.10  & 88.56   & 62.34 & 51.70   &  59.69 &  55.40 \\
		BoW &  BoW  & Yes (S) & MLP & TF-IDF & 65.99 & 64.14 & 65.05 & 48.09 & \textbf{98.03} & \textbf{63.17} & 50.83 & 65.22 & \textbf{57.13} \\

		\hline
		
	\end{tabular}%
}
	\caption{Consistency performance (average of 3 runs) from our models 
	 (see \cref{sec:experiments:components} for a detailed explanation of the columns). %
	We indicate existing credibility models with $\clubsuit$ \cite{popatDeclare} and $\spadesuit$ \cite{rashkinFake}. BoW refers to bag of GLoVE embeddings.}
	\label{tab:consistencyPerformance}%
\end{table*}%

\begin{table*}[t]
	\centering
	\resizebox{0.99\textwidth}{!}{
	\begin{tabular}{|c|c|c|c|c||c|c|c||c|c|c||c|c|c|}
		\hline
			\textbf{$f_{belief}$} & \textbf{$f_{evidence}$} & \textbf{Comb.} & \textbf{Feat.} & \textbf{Sparse} & \multicolumn{3}{c||}{\textbf{TACRED-KG}} & \multicolumn{3}{c||}{\textbf{TAC-2017}} & \multicolumn{3}{c|}{\textbf{TAC-2015}} \\
		\hline
			& & & &  & \textbf{Macro} & \textbf{Micro} & \textbf{MRR} & \textbf{Macro} & \textbf{Micro} & \textbf{MRR} & \textbf{Macro} & \textbf{Micro} & \textbf{MRR} \\
		\hline		\hline

		None & None & No & None &   Binary & 2.16 & 41.65 & 0.83 & 44.86 & 53.10 & 0.83 & 22.78 & 16.50 & 0.19 \\
                
                None & None & No & None &  TF-IDF & \textbf{14.50} & 43.48 & \textbf{0.83}  & 75.49 & 76.80 & 0.76  & \textbf{76.35} & 77.57 & 0.76 \\
		 
		\hline
		\hline
		None & $\spadesuit$  LSTM  & No & MLP & No & 1.87  &  \textbf{78.56} & 0.82  & 3.05  & 33.04 & 0.53  & 1.46  & 61.30 & 0.68 \\
		\hline
		BoW & $\clubsuit$  Bi-LSTM  & Yes (W) & MLP & No & 1.24  & 52.39 & 0.8  & 1.04  & 32.02  & 0.43  & 1.46  & 61.30 & 0.66 \\
		\hline		\hline
		BERT & BERT  & Yes (S) & MLP & TF-IDF & 4.10 & 7.72 & 0.28 & 72.17 & 81.85 & 0.89 & 54.91 & 58.61 & 0.69\\
		BoW & BoW  & Yes (S) & MLP & TF-IDF & 7.22 & 64.43 & 0.74 & \textbf{76.39} & \textbf{85.33} & \textbf{0.91} & 65.76 & \textbf{78.02} & \textbf{0.86} \\
		
		\hline
	\end{tabular}%
}
	\caption{Repair Performance (averaged over 3 runs) of models with abbreviations as in Table \ref{tab:consistencyPerformance}.}

	\label{tab:repairPerformance}%
\end{table*}%

\subsection{Components}
\label{sec:experiments:components}

We  evaluated the effect of each of the four major components mentioned below. %
We used Glove \cite{pennington2014glove} as pre-trained word embeddings, except for BERT models, where we used the uncased base model~\cite{devlin2019bert}. %


\textbf{Representations (Rep.)}: We evaluated three ways to represent beliefs and provenance text (compute $f_{\textit{belief}}$ and $f_{\textit{evidence}}$): \textit{Bag-of-Words (BoW) embedding} which is the average of Glove embeddings, the final output from the LSTM and Bi-LSTM models, and the BERT representation output. 
While an average of embeddings may seem simple, this approach has empirically performed well on other tasks compared to more complicated models~\cite{iyyer2015deep}. %


\textbf{Combining belief \& provenance (Comb.)}: 
When beliefs and provenance are used, we considered similarity as sentence-level attention (``Yes (S)'') as well as word-level attention (``Yes (W)''). 


\textbf{Feature Learning (Feat.)}: In our primary experiments to do further feature learning we used a three layer multi-layer perceptron (``MLP'') to do further feature learning. %
We indicate no further feature learning with a value of ``None.'' %

\textbf{``Blob'' Sparse Connection (``Sparse'')}: If used, we set $f_{\textit{blob}}$ to compute either a TF-IDF or binary-lexical vector based on the \textit{blob} (concatenation of all sentences for a belief). This computed representation skips the feature learning component and is provided directly to the classifier.

\subsection{Results} 
\label{sec:results}

\label{sec:result}

The overall test results across our three datasets are shown in \cref{tab:consistencyPerformance} for the consistency task and \cref{tab:repairPerformance} for the repair task. Each of the selected models was, prior to evaluation on the test set, chosen due to its performance on development data. The results are averaged across three runs.

\subsubsection{Can Credibility Models be Used?}
\label{sec:credibilitymodel}
We first examine and compare our proposed framework against two different strong performing credibility models. These external methods are our baselines and we indicate them in Tables \ref{tab:consistencyPerformance} and \ref{tab:repairPerformance} by ``$\clubsuit$'' \cite{popatDeclare} and ``$\spadesuit$'' \cite{rashkinFake}. We find they both perform poorly compared to other models, indicating that while both tasks learn similar functions 
the credibility models cannot be used ``as-is'' for consistency. 
This highlights the fact that the consistency task is sufficiently different from the existing credibility task.

Moreover, in examining whether credibility models transfer to the repair task, word level attention with a Bi-LSTM sentence encoder, as in DeClarE \cite[$\clubsuit$]{popatDeclare}, performs poorly in the repair task too (with one exception on TACRED-KG).  %
These results highlight differences in the credibility vs. consistency tasks, and the applicability of existing credibility models to both consistency and repair, suggesting that a dedicated framework and study such as ours is needed. %

\begin{table}[t]
	\centering
        \resizebox{.99\columnwidth}{!}{
		\begin{tabular}{|c|c|c||c|c|c||c|c|c|}
			\hline
			$f_{belief}$ and & \multirow{2}{*}{\textbf{Comb.}} &  \multirow{2}{*}{\textbf{Sparse}} &  \multicolumn{3}{c||}{\textbf{Consistency}} & \multicolumn{3}{c|}{\textbf{Repair}} \\
			\cline{4-9}		
			 $f_{evidence}$ & & & \textbf{P} & \textbf{R} & \textbf{F1} & \textbf{Macro} & \textbf{Micro} & \textbf{MRR}\\
			\hline
			BoW &   No  & No   & 12.01 & 33.33 & 17.65 & 0.92 & 22.08 & 0.38 \\ \hline
			BoW &   Yes (S) & No & 12.01 & 33.33 & 17.65 & 0.89 & 21.16 & 0.34 \\ \hline
			BoW &   No  & TF-IDF & 47.98 & 90.75 & 62.77 & 75.71 & 85.24 & 0.90 \\ \hline
			BoW &   Yes (S)  & TF-IDF & 48.09 &  \textbf{92.03} &  63.17 & \textbf{76.39} & \textbf{85.33} & \textbf{0.91} \\ \hline
			
			Bi-LSTM   & Yes (S) & TF-IDF & \textbf{59} & 87.71 & \textbf{70.53} & 75.76 & 83.86 & 0.89 \\ \hline
            BERT & Yes (S) & TF-IDF & 48.11 & 91.47 & 63.06 &  76.30& 85.25& 0.91 \\ \hline
			
			\hline
		\end{tabular}%

	}
	\caption{Consistency and repair performance ablation study, averaged over three runs. "Comb." is belief and provenance combination, and "Skip" is the use of skip connection. All use an MLP for feature learning. For space, we only consider TAC 2017 in these experiments.}
	\label{tab:consistencyRepairPerformanceAblation}%
\end{table}%

\subsubsection{What Representations are Effective?}
\label{sec:features}

\textit{Consistency}: Both sentence attention and a TF-IDF sparse connection improve the overall F1 of our framework's embedding-based models. 
We noticed that precision and recall vary across the datasets due to their different characteristics. This can be seen with the two methods that rely only on the lexically-based sparse connections (the first two rows of \cref{tab:consistencyPerformance}): while performance was strong on TACRED-KG consistency, it was quite poor on TAC 2015 and 2017. These latter two datasets have more provenance sentences per belief, and make fewer assumptions about what must be contained in the provenance. %
Together, this results in greater lexical variety, which suggests that while non-neural lexical-based consistency approaches can be effective in settings with limited provenance, stronger approaches are needed for greater and more diverse provenance. %
Learning refined embeddings (rows 5 and 6) suggests that these pre-trained models are helpful in the task. 
BERT benefits from the less noisy provenance in TACRED-KG. However, similar or slightly better performance is achieved when simple word embeddings are used, especially for TAC 2015/2017, highlighting the difficulty of the consistency task with noisier provenance.


\begin{figure*}
	\centering
	\resizebox{\textwidth}{!}{
	\begin{tabular}{|p{16cm}|} 
		\hline
		\textbf{Belief}: {\it  Marty Walsh; org:city\_of\_headquarters;} {\it  Neighborhood House Charter School} \\
		\textbf{Summary}: (\ding{51}, {\it  fixed})  \\
		\textbf{Human(C)}: No; \textbf{Predicted(C)}: No; \textbf{Human(R)}: org:founded\_by; \textbf{Predicted(R)}: org:founded\_by\\
		\textbf{Provenance}: Walsh was a founding board member of Dorchester's Neighborhood House Charter School, and makes clear that he would support lifting the cap on charters in the city, something that hardly wins him the favor of the Boston Teachers Union. \\
                \hline


		\textbf{Belief}: {\it  Alan M. Dershowitz; per:title; professor} \\ 	 
		\textbf{Summary}: (\ding{55}, {\it incorrect\_fixed})  \\
		\textbf{Human(C)}: Yes; \textbf{Predicted(C)}: No; \textbf{Human(R)}: per:title; \textbf{Predicted(R)}: per:religion\\
		\textbf{Provenance}: Harvard Law professor Alan Dershowitz said Sunday that the Obama administration was naive and had possibly made a "cataclysmic error of gigantic proportions" in its deal to ease sanctions on Iran in exchange for an opening up of the Islamic Republic s nuclear program. \\
                \hline               
	\end{tabular}%
}

	\caption{Examples of our model's predictions on the TAC 2015 datasets. Human: gold standard label, Predicted: our model's label, C: Consistency, R: Repair, Human(C): Human Consistency label, and Predicted(C): Predicted consistency label. Similarly for repair. Summary indicates overall prediction analysis of example. (\ding{51}, fixed) means consistency correctly predicted and incorrect belief was fixed.}
	\label{fig:predictionStudy}%
\end{figure*}%

\textit{Repair}: Perhaps surprisingly, an embedding model with a TF-IDF sparse connection yielded good performance. The sparse-based lexical features are most influential, as evident from when just TF-IDF or binary lexical features are used. Looking across the three datasets, we notice that a TF-IDF only model provides a surprisingly strong baseline, outperforming the existing credibility models in almost all cases. 
Using BoW embedding with sentence attention, MLP feature learning, and a TF-IDF sparse connection, we can surpass a sparse-only TF-IDF approach. The BERT-based representation, fine-tuned or not, performed nearly equally to a BoW embedding on the repair task, indicating both the effectiveness of its pre-trained model and highlighting the difficulty of this repair task.


\subsubsection{How Helpful Is Sequential Modeling?}
\label{sec:sequential}
As indicated by \newcite{zhang2017position}, the sentences in TACRED and TAC are long. Consistency and repair models must be able to handle that. Note that BoW representation methods do not consider word order, while LSTM, Bi-LSTM and BERT embeddings do. From Tables \ref{tab:consistencyPerformance} and \ref{tab:repairPerformance}, we see that TF-IDF sparse features and a sentence level combination of the belief and provenance give the best performance on both tasks when using a BoW representation, as compared to an LSTM, Bi-LSTM with word attention, and BERT. This indicates that for consistency and repair, \textbf{unordered lexical features can be sufficient to get better performance}. 

We further examine this in \cref{tab:consistencyRepairPerformanceAblation}, where due to space we focus on TAC 2017. Notice that while sequence-based encodings can improve some aspects (e.g., precision and F1 for consistency), there are not across-the-board improvements. We experimented with replacing the BoW embedding with a sentence-level Bi-LSTM representation. A Bi-LSTM representation with just attention and TF-IDF sparse features gives better consistency precision and F1 compared to BoW embedding approaches. However, the Bi-LSTM results in overall lower performance for repair. While the differences are not very large, they indicate that \textbf{simple methods can outperform, or perform competitively with, sequential and autoencoding methods}. 

\subsubsection{Relation Repair vs. Re-Extraction}
\label{sec:relationExtraction}
While the repair task \textit{can} be viewed as relation re-extraction, we examine the implications of this. %
Tables. \ref{tab:consistencyPerformance} and \ref{tab:repairPerformance} show a large performance drop for TACRED-KG vs. TAC 2015/2017.  First, TACRED was created from a TAC dataset and modified and augmented by crowd-sourced workers. When the belief was found with abstract or generalized provenance, workers were shown a set of sentences containing the subject-object pairs and asked to pick the representative sentence which was most specific.  Second, each sentence is guaranteed to include the subject and object mentions, which is not always true for TAC 2015 and 2017, where a significant number of TAC provenance sentences were missing one or both the subject and object mentions. 
This highlights some of the differences in the core assumptions made in the construction of a relation extraction dataset.

\subsection{Prediction Error Analysis}
\Cref{fig:predictionStudy} demonstrates our framework's performance on some examples from TAC 2015. %
The first example describes the case where the belief was consistent with the provenance information and there was no recommendation of an alternate relation. 
Depending on the provenance the fix may not be appropriate, as in the second example of per:title vs. per:religion where we believe an indicative word like ``Islamic'' influenced the repair prediction.

\subsection{Ablation Study}
\label{sec:ablationstudy}

Our results show the strength of attention with lexical features. 
We further examine the impact of lexical features, using the first four rows of \cref{tab:consistencyRepairPerformanceAblation}. %

\textbf{Lexical Impact on Consistency.} 
From the first row of \cref{tab:consistencyRepairPerformanceAblation}, we see BoW embedding for both the belief and provenance results in low precision and recall. While adding attention does not help, using TF-IDF sparse features drastically improves performance. Meanwhile, removing sentence-based attention only has a small impact on performance. All together this indicates the provenance found by the IE system is \textit{more lexically systematic}. 

\textbf{Lexical Impact on Repair.} A similar trend is seen for the repair task: our combined representation with TF-IDF is better than relying only on embeddings. Combining belief and provenance sentences gets slightly better micro overall compared to macro. This affects the MRR score too. However, the best performance is achieved when all components are combined. 

\section{Related Studies}
There has been research on determining the consistency of beliefs  using either schemas or ensembles, but none that are language-based, do not require access to IE system details, or attempt to repair inconsistent facts. Our work addresses all these. 



\textbf{Schema and Ensemble Based approaches}:
Previous work by \citet{KGEval} and \citet{pujara2013knowledge} determined the consistency of the extracted belief using a schema as the side information and coupling constraints to satisfy the schema's axioms. Rather than applying schemas, \newcite{yu2014wisdom} proposed an unsupervised method applying linguistic features to filter credible vs. non-credible belief. However, it required access to multiple IE systems with different configuration settings that extracted information from the same text corpus. Viswanathan et al. \shortcite{viswanathan2015stacked} used a supervised approach to build a classifier from the confidence scores produced by multiple IE systems for the same belief. These are not \textit{standalone} systems, as they assume the availability of multiple IE systems.


%

\textbf{Language based approaches}:
The FEVER \cite{thorne2018fever} fact-checking study proposes a framework for credibility task and performs provenance-based classification without attempting to repair errors. 
This task has inspired a number of efforts~\citep[i.a.,]{yin-roth-2018-twowingos}, including \citet{ma-etal-2019-sentence} who tackle a problem similar to our consistency.
\citet{10.1162/tacl_a_00454} outlines additional language-based approaches for consistency prediction (they term it ``verdict prediction''). %
However, a crucial difference is that we aim to operate on KG tuple outputs as the belief (not sentences).

Overall, our study differs from previous ones in two important ways.
(1) We address the problem of determining consistency and potential corrections without access to an underlying semantic schema. %
(2) Our standalone approach treats the underlying IE systems as \textit{blackboxes} and requires no access to the original IE systems or detailed system output containing confidence scores. 

\section{Conclusions}

We propose a task of refining the beliefs produced by a blackbox IE system that provides no access to or knowledge of its internal workings. First we analyze the types of errors made. Then we propose two subtasks: determining the consistency of an extracted belief and its provenance text, and suggesting a repair to fix the belief. We present a modular framework that can use a variety of representation, and learning techniques, and subsumes prior work. This framework provides effective techniques for the consistency and repair tasks. 



{
\section*{Acknowledgements}
We would also like to thank the anonymous reviewers for their comments, questions, and suggestions. %
This material is based in part upon work supported by the National Science Foundation under Grant Nos. IIS-1940931, IIS-2024878, and DGE-2114892. %
Some experiments were conducted on the UMBC HPCF, supported by the National Science Foundation under Grant No. CNS-1920079.%
This material is also based on research that is in part supported by the Army Research Laboratory, Grant No. W911NF2120076, and by the Air Force Research Laboratory (AFRL), DARPA, for the KAIROS program under agreement number FA8750-19-2-1003. The U.S.Government is authorized to reproduce and distribute reprints for Governmental purposes notwithstanding any copyright notation thereon. The views and conclusions contained herein are those of the authors and should not be interpreted as necessarily representing the official policies or endorsements, either express or implied, of the Air Force Research Laboratory (AFRL), DARPA, or the U.S. Government. %
}

\nocite{padiaPhd}

\bibliography{anthology,custom}
\bibliographystyle{acl_natbib}


\end{document}